%% file: main.tex
\documentclass[journal]{IEEEtran}

\usepackage{graphicx}
\usepackage{amsmath}
\usepackage{amssymb}
\usepackage[dvipsnames]{xcolor}
\usepackage{xspace}
\usepackage{graphbox}
\usepackage{tabu}
\usepackage{ragged2e}

\usepackage{url} 
\usepackage{import}
\usepackage{multirow}
\usepackage{bm}
\usepackage{makecell}
\usepackage{booktabs}
\newcommand\Tstrut{\rule{0pt}{2.2ex}}     
\newcommand\Bstrut{\rule[-1ex]{0pt}{0pt}} 
\newcommand{\TBstrut}{\Tstrut\Bstrut}

\makeatletter
\DeclareRobustCommand\onedot{\futurelet\@let@token\@onedot}
\def\@onedot{\ifx\@let@token.\else.\null\fi\xspace}

\def\eg{\emph{e.g}\onedot} 
\def\ie{\emph{i.e}\onedot}

\def\etal{\emph{et al}\onedot}
\makeatother

\RequirePackage{enumitem}
\setlist[itemize]{nosep}
\renewcommand{\paragraph}[1]{\vspace{0.2em}\noindent \textbf{#1 \hspace{0.2em}}}

\def\uvmodel{Multi-reference UV-map Completion Network\xspace}
\def\uvmodelAbbr{MUC-Net\xspace}
\def\uvattn{Frame-wise Attention\xspace}
\def\uvattnAbbr{FA\xspace}
\def\refine{Face Video Refinement Network\xspace}
\def\refineAbbr{FVR-Net\xspace}

\usepackage{cite}

\ifCLASSINFOpdf
\else
\fi

\ifCLASSOPTIONcompsoc
 \usepackage[caption=false,font=normalsize,labelfont=sf,textfont=sf]{subfig}
\else
 \usepackage[caption=false,font=footnotesize]{subfig}
\fi

\ifCLASSOPTIONcaptionsoff
 \usepackage[nomarkers]{endfloat}
\let\MYoriglatexcaption\caption
\renewcommand{\caption}[2][\relax]{\MYoriglatexcaption[#2]{#2}}
\fi

\usepackage{url}

\begin{document}

\title{Deep Face Video Inpainting\\ via UV Mapping}

\author{Wenqi~Yang,
        Zhenfang~Chen,
        Chaofeng~Chen,
        Guanying~Chen,
        and~Kwan-Yee~K.~Wong,~\IEEEmembership{Senior~Member,~IEEE}
\thanks{Wenqi Yang and Kwan-Yee K. Wong are with the Department of
Computer Science, The University of Hong Kong, Pokfulam Road, Hong
Kong. (E-mail: [wqyang, kykwong]@cs.hku.hk)}
\thanks{Zhenfang Chen is with MIT-IBM Watson AI Lab, Cambridge, MA, USA, 02142 (E-mail: chenzhenfang2013@gmail.com).}
\thanks{Chaofeng Chen is with Nanyang Technological University, Singapore (E-mail: chaofenghust@gmail.com).}
\thanks{Guanying Chen is with FNii and SSE, The Chinese University of Hong Kong, Shenzhen, China (E-mail: chenguanying@cuhk.edu.cn).}%
}

\maketitle

\import{./text}{0_abstract.tex}
\IEEEpeerreviewmaketitle

\import{./text}{1_introduction.tex}

\import{./text}{2_related_work.tex}

\import{./text}{3_method.tex}

\import{./text}{4_experiments.tex}

\import{./text}{5_conclusion.tex}

\section*{Acknowledgment}
Guanying Chen is supported in part by NSFC with Grant No. 62293482 and 62202409, and the Guangdong Provincial Key Laboratory of Future Networks of Intelligence (Grant No. 2022B1212010001)

\ifCLASSOPTIONcaptionsoff
  \newpage
\fi

\bibliographystyle{IEEEtran}

\bibliography{IEEEabrv,egbib}

\end{document}

%% file: text/0_abstract.tex

\begin{abstract}

This paper addresses the problem of face video inpainting. Existing video inpainting methods target primarily at natural scenes with repetitive patterns. They do not make use of any prior knowledge of the face to help retrieve correspondences for the corrupted face. They therefore only achieve sub-optimal results, particularly for faces under large pose and expression variations where face components appear very differently across frames. In this paper, we propose a two-stage deep learning method for face video inpainting. We employ 3DMM as our 3D face prior to transform a face between the image space and the UV (texture) space. In Stage~I, we perform face inpainting in the UV space. This helps to largely remove the influence of face poses and expressions and makes the learning task much easier with well aligned face features. We introduce a frame-wise attention module to fully exploit correspondences in neighboring frames to assist the inpainting task. In Stage~II, we transform the inpainted face regions back to the image space and perform face video refinement that inpaints any background regions not covered in Stage~I and also refines the inpainted face regions. Extensive experiments have been carried out which show our method can significantly outperform methods based merely on 2D information, especially for faces under large pose and expression variations. Project page: \url{https://ywq.github.io/FVIP}.

\end{abstract}

%% file: text/1_introduction.tex

\section{Introduction} \label{sec:intro}

\IEEEPARstart{F}{ace} video inpainting targets at restoring corrupted or occluded regions of faces in videos. It is an important research topic in computer vision and has many practical applications such as video overlay removal~\cite{3dconv-vd} and partially occluded face recognition in surveillance videos~\cite{mathai2019does}. Note that faces in videos often exhibit diverse poses and expressions. 
This makes face video inpainting a challenging task.

\import{imgs/}{fig-intro.tex}

Correspondences between frames serve as crucial clues in video inpainting for retrieving missing information from neighboring frames and ensuring temporal consistency. 
Existing video inpainting methods mainly focus on restoring the backgrounds of natural scenes which are mostly stationary and consist of repetitive patterns.
They typically  fill missing regions by copying and propagating similar patterns or textures from other regions \cite{tempo-coherent-vd, newson2014video, 7112116, 4060949}. 
However, directly referring to other frames often results in improper contents when elements in a video move around and change their appearances. Hence, these methods are only capable of tackling narrow masks and static background.
Recently, a number of learning-based methods have been proposed \cite{3dconv-vd, 3dconv-2-vd, frame-recurrent-vd, deep-vd-inpaint, copy-paste-vd, short-long-vd, vd-temp-spatial, flow-vd}. These methods successfully learn domain knowledge from an enormous number of training samples and can generate proper content for large missing regions. 
Most of them are based on spatial-temporal attention or assisted by optical flow to learn the correspondences across frames, and are suitable for natural scenes with simple motions.
For face videos, however, the appearance of a face can vary a lot under different poses and expressions.  These methods have difficulties in finding proper reference in neighboring frames to restore reasonable contents for faces. They often fail to generate visually plausible face structures when no reference can be found in neighboring frames due to their lack of face prior knowledge. Hence, they cannot guarantee recovering proper faces in the videos.

Owing to the prior knowledge of the 3D face structure, human can interpret, recognize, or even ``reconstruct'' a corrupted face image with relative ease. For instance, human can recognize faces in low quality videos under diverse viewpoints and partial occlusions, as well as under different face poses and expressions. 
Inspired by this, we propose to exploit 3D face prior for face video inpainting. 
In this paper, we employ an expressive 3D face model as our 3D face prior. By fitting this 3D face model to the video frames, we can transform the face from the image space to the UV (texture) space and vice versa. 
Note that faces in the UV space represent unwarped face textures which are well aligned. This helps to remove the influence of poses and expressions, and makes the learning of face structure much easier. 
Besides, the well alignment and symmetry of the face features in the UV space also make it trivial to locate correspondences in neighboring frames which provide rich information for face video inpainting. Based on these observations, we propose to carry out face video inpainting in the UV space (see~Fig.~\ref{fig-intro}). We introduce a \uvmodel (\uvmodelAbbr) with a novel \uvattn (\uvattnAbbr) module to perform reference-based face completion in the UV space.

Our proposed method is a two-stage approach. As a pre-processing step, we fit the 3D Morphable Model (3DMM) \cite{3dmm} to every frame of the face video. We use the estimated model parameters to transform the face between the image space and the UV space in the two core stages. In Stage~I, namely UV-map completion stage, we first transform the face to the UV space and carry out UV-map completion using our proposed \uvmodelAbbr. Our \uvattnAbbr module is designed specifically to take full advantage of the well-aligned face features in the UV space to find proper correspondences in neighboring frames in an efficient and effective manner. In Stage~II, namely face video refinement stage, we transform the inpainted UV-map back to the image space and perform face video refinement using our proposed \refine (\refineAbbr). \refineAbbr inpaints any background regions not covered in Stage~I and at the same time refines the inpainted face regions.

In contrast to other methods, our method ensures the plausibility of face structure through the use of 3D face prior. Our method is more robust for faces under large pose and expression variations, and can better exploit correspondences in neighboring frames. Our key contributions include:
\renewcommand{\labelitemi}{$\bullet$}
\begin{itemize}
    \item To the best of our knowledge, we are the first to perform face video inpainting via the UV space. Thanks to the well alignment and symmetry of the face features in the UV space, our \uvmodelAbbr can robustly restore the missing face regions with plausible face structures and textures.
    \item We propose a novel \uvattn (\uvattnAbbr) module that can take full advantage of the well aligned face features in the UV space to find proper correspondences efficiently in neighboring frames to assist face inpainting.
    \item Our method achieves state-of-the-art performance in face video inpainting, especially for the challenging cases with large face pose and expression variations. Comprehensive experiments demonstrate the effectiveness and robustness of our method.
\end{itemize}

%% file: imgs/fig-intro.tex
\begin{figure}[ht]
\begin{center}
   \includegraphics[width=\linewidth]{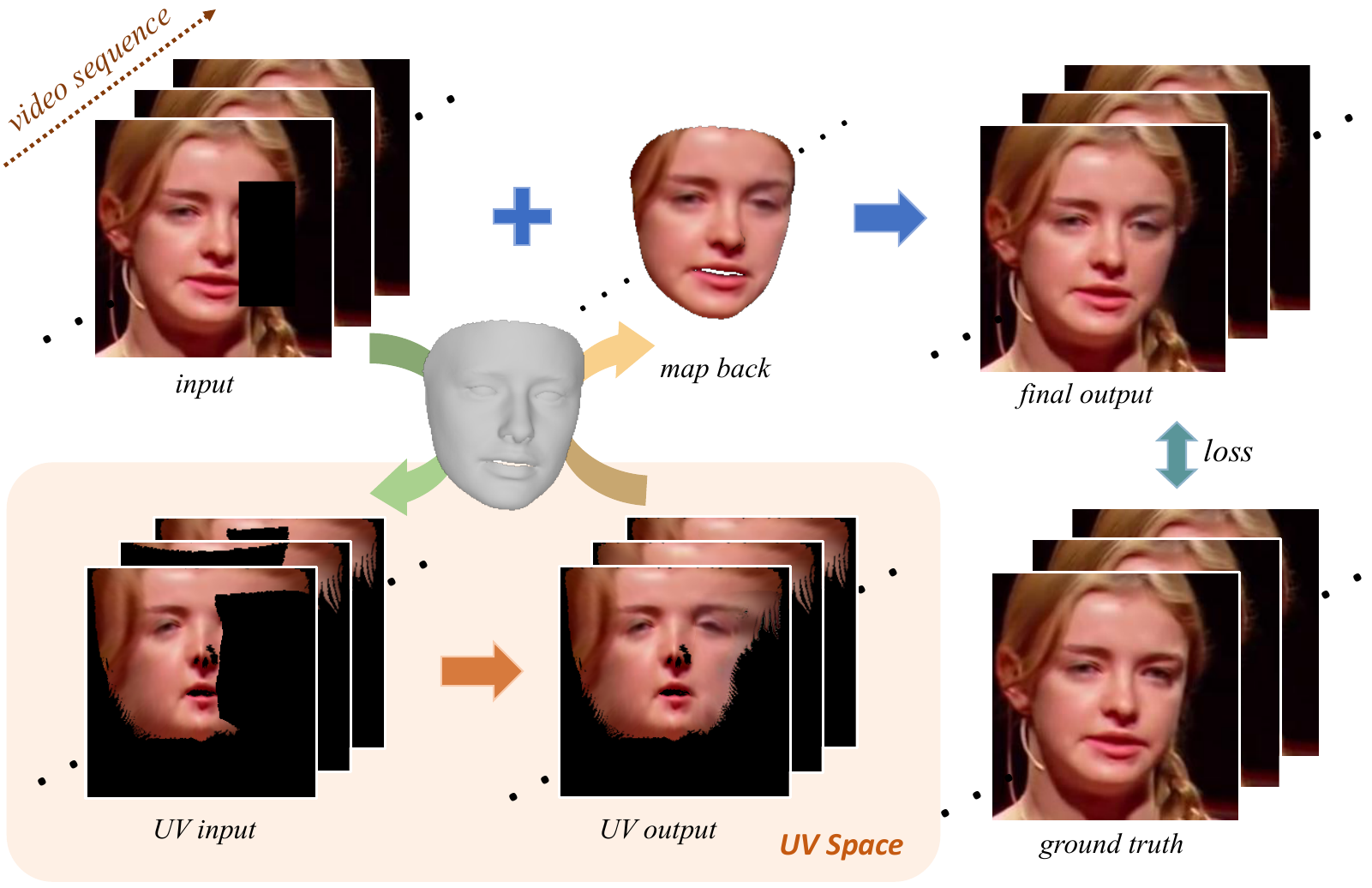}
\end{center}
   \caption{Given a face video, it is preferable to learn the face texture restoration regardless of face pose and expression variations. In our proposed method, we first utilize 3D face prior (3DMM) to transform the input faces into the UV space. We then perform face inpainting in the UV space where face textures are well aligned and easy for correspondence retrieval. The restored face will then be mapped back to the image space, followed by final refinement as well as inpainting for the non-face regions to produce the final outputs.}
\label{fig-intro}
\end{figure}

%% file: text/2_related_work.tex

\section{Related Work}

\subsection{Face Image Inpainting}
Traditional image inpainting methods fill the missing regions by progressively propagating pixels from the neighboring regions \cite{bertalmio2000image-img,ballester2001filling, structure-texture-2003-img} or by iteratively searching for matching patches \cite{criminisi2004region, patchmatch-2009-img, exemplar-2003, efros2001image, hays2007scene}. 
These conventional methods are capable of handling cases with stationary textures and relatively small masked regions, but fail when textures and structures are non-repetitive.
Some studies~\cite{hwang2003reconstruction,face-inpainting-2004-img} exploit specific face domain knowledge for face image inpainting. Constrained by the representation ability of their models, however, they can only restore specific face regions in frontal faces.

Recently, learning-based methods \cite{global-local-inpaint-img, context-encoder-img, yang2017high, yan2018shiftnet, wang2021image, xie2019image, zhang2018semantic, shao2020multi, jia2010reflection, wang2022dual, ni2022n} have been proposed to perform inpainting by learning from large image datasets. These methods are more robust and expressive than the traditional non-learning-based methods. 
Some of them focus on improving the structure and texture coherence by introducing additional guidance such as edges~\cite{edgeconnect-img}, segmentation~\cite{segmentation-img}, structure flow~\cite{structureflow-img}, and foreground contour~\cite{foreground-img}. 
Others \cite{pyramid-img, liu2019coherent, sagong2019pepsi} explore feature matching between masked regions and known regions in the semantic space by proposing new modules such as contextual attention~\cite{generative-inpaint-img}.
There are also inpainting works that focus on progressively filling in the masked regions from the their boundaries and treating masked regions (usually input as 0 value) and non-masked regions separately, such as partial convolution~\cite{pconv-img} and gated convolution~\cite{gated-img}. Partial convolution~\cite{pconv-img} updates a binary mask regularly, while gated convolution~\cite{gated-img} adopts a learnable soft valued mask.

In response to the need of face related applications, a number of methods \cite{lahiri2018improved-img-face, hwang2003reconstruction, face-inpainting-2004-img} have been proposed specifically for the face inpainting task. 
To stabilize the restored face structures, some of them introduce additional guidance such as landmarks~\cite{lafin-img-face,zhang2020domain-img-face} and face parsing~\cite{li2017generative, song2019geometry-img-face} into the pipeline to serve as intermediate output or loss term. 
Others propose to make use of additional inputs such as reference images~\cite{identity-bmvc-2018-img-face} from the same person to preserve identities, 
or colorized sketches~\cite{sc-fegan-img-face,faceshop} to perform interactive face editing (modify the shape / color of the given face). 
Li \etal~\cite{li2020learning-img-face} propose to utilize face symmetry by performing illumination-aware feature warping from flipped images. However, human face may not look strictly symmetric under large pose variations. 
Another batches of works~\cite{li2018gfrnet,li2020asffnet,li2020dfdnet,ChenPSFRGAN,wang2021towards} focus on blind face image restoration without masks indicating corrupted regions.
Based on observations, most of the face image inpainting solutions perform unsatisfactorily under large pose or expression variations. This is due to their lack of 3D face priors to help understand and restore face structures from 2D images. 
Furthermore, these image-based methods can only achieve sub-optimal results in face video inpainting as they do not exploit information provided by correspondences in neighboring frames.

\import{imgs/}{fig-framework.tex}

\subsection{Video Inpainting}\label{video-related}
Different from image inpainting, video inpainting takes a sequence of frames as input and restores the missing regions based on both spatial and temporal information. Compared with image-based methods, video inpainting methods explore correspondences between frames as crucial clues to retrieve missing information from neighboring frames to ensure temporal consistency.
Traditional video inpainting methods \cite{tempo-coherent-vd, newson2014video, patwardhan2005video-vd-trad, wexler2007space-vd-trad} typically perform patch-based or optical-flow-based optimizations which require heavy computations.
They are capable of generating plausible contents for general videos consisting of stationary background with repetitive patterns and consistent textures. However, they may fail miserably when structures and textures are complicated and their appearances vary largely across frames.
Boosted by deep learning techniques, learning-based methods have been proposed to explore solutions for better utilizing spatial and temporal information by introducing flow-warping \cite{flow-vd, gao2020flow-vd, copy-paste-vd, deep-vd-inpaint, frame-recurrent-vd, zou2021progressive}, cross-frame attention~\cite{short-long-vd, onion-vd}, and 3D convolution \cite{vd-temp-spatial, 3dconv-vd, 3dconv-2-vd}. 
Optical flow is often adopted as an intermediate guidance \cite{flow-vd, gao2020flow-vd} or used to warp frames into alignment \cite{zou2021progressive}. This facilitates the calculation of warping loss \cite{deep-vd-inpaint}. 

The above methods mainly retrieve correspondences by searching for similar patches or making the patterns aligned based on flow-warping. However, human face can appear very differently under large pose and expression variations. 
This makes it more difficult to find proper reference from neighboring frames due to the large appearance variations.
There is also a video inpainting work \cite{img2vd-vd-face} focusing on face re-identification.
They target at the restoration of the de-identified face videos by taking the original face landmarks as input. The mask is designed to cover all the key face components for all the input frames while the background is preserved. Under this setting, no reference is available from other frames to recover the masked regions. They instead focus on predicting consistent identity for all the frames from the given landmarks.
In this paper, we aim at efficiently retrieving proper correspondences from neighboring frames for face video inpainting. We exploit an effective way to transform face textures into a well-aligned space which greatly facilitates both correspondence learning and feature restoration.

\subsection{3D Face Prior}
Human domain knowledge has become a powerful tool in numerous tasks owing to the learnable human prior (e.g., body structure~\cite{wu2019deep} and face prior~\cite{de-occlu-3dmm-img-face}). 
In this paper, we focus on face prior assisted face video inpainting.
Commonly used face priors include face parsing, face landmarks, and face model \cite{Zeng_2019_ICCV, chaudhuri2020personalized}. In particular, 3D face morphable model (3DMM) \cite{3dmm, accu-face-recons-19, egger20203d, sariyanidi2020inequality, lin2020towards, egger20203d} has achieved stable and excellent performance in face reconstruction, and has been wildly adopted in face related works such as face recognition\cite{uv-gan}, face frontalization~\cite{gecer2021ostec}, face editing \cite{cao2020task}, makeup transfer\cite{nguyen2021lipstick}, face reenactment\cite{xu2020deep}, face super-resolution~\cite{hu2020face}, face deblurring~\cite{ren2019face}, and animation \cite{lee1997model}.
The impressive results of these works well demonstrate the advantages of embedding 3D face prior into face-related tasks.

Among works that utilize 3D face prior, UV-GAN~\cite{uv-gan} is closely related to our work. UV-GAN also utilizes face model and UV map to recover face regions. However, their motivation and contributions are different from ours. UV-GAN is proposed to reconstruct face models and synthesize novel views to enlarge the diversity of poses for pose-invariant face recognition. They exploit UV textures and leverage symmetry of the face to recover self-occluded regions in the fitted model. They only deal with single images.
In this paper, we target at robust face video inpainting by making use of 3D face prior to facilitate both face structure learning and correspondence finding from the well-aligned feature maps in the UV space.

%% file: imgs/fig-framework.tex
\begin{figure*}
\begin{center}
\includegraphics[width=0.99\linewidth]{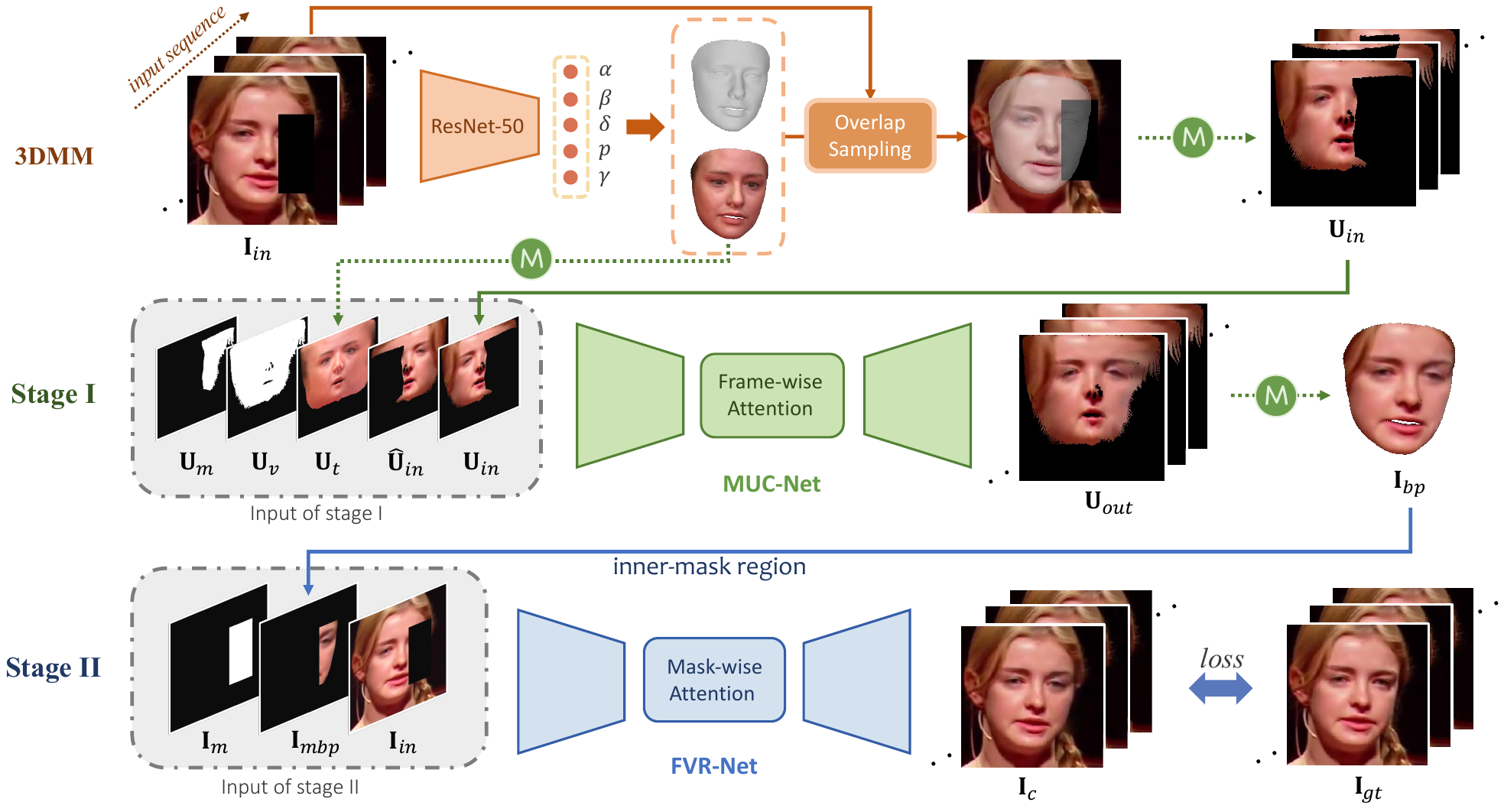}
\end{center}
   \caption{The overall pipeline of our method, where 
  ``\raisebox{0pt}{\textcircled{\raisebox{-0.7pt} {\scriptsize{M}}}}'' represents UV mapping.
  Our framework consists of two stages, namely (I) UV map completion and (II) Face video refinement. Given a face video sequence, we first  utilize 3DMM as our face prior to predict the face structures by model regression. In Stage I, we transform the face texture from image space to UV space via the fitted 3DMM, and perform UV map completion to get preliminary results. In Stage II, we take the mapped-back face together with the corrupted face to generate the final output, where both face regions and background (non-face regions) will be refined or inpainted.
   }
\label{fig-framework}
\end{figure*}

%% file: text/3_method.tex
\section{Method}

\subsection{Overview}
As briefly introduced in Sec.~\ref{sec:intro}, our method is a two-stage approach. Fig.~\ref{fig-framework} shows an overview of our proposed pipeline. Given a face video, we first fit 3DMM to every frame to obtain per-frame shape, texture, and pose parameters. The shape and pose parameters are used for transforming the face between the image space and UV space, while the texture parameters are used to generate synthesized texture to provide auxiliary information for the inpainting task. In Stage~I, we first transform the face from the image space to the well-aligned UV space and use our proposed \uvmodelAbbr to perform UV-map completion.  \uvattnAbbr module is proposed for  \uvmodelAbbr to facilitate the correspondence retrieval across UV texture frames.
In Stage~II, we transform the completed UV-map in Stage~I back to the image space, and use our proposed \refineAbbr to inpaint any background (non-face) regions not covered in Stage~I as well as refine the inpainted face regions.

In Sec.~\ref{sec:3DMM}, we will first give a brief review of 3DMM which serves as our 3D face prior and facilitates the transformation of faces between the image space and the UV space. We will then describe the details of our \uvmodelAbbr and \uvattnAbbr module for UV-map completion in \ref{sec:Stage-I}. Details of face video refinement using our \refineAbbr will be covered in Sec.~\ref{sec:Stage-II} .

\subsection{3D Face Prior}\label{sec:3DMM}

\subsubsection{Face Reconstruction}
In this work, we employ 3DMM \cite{3dmm} as our 3D face prior. We adopt the method proposed by Deng \etal~\cite{accu-face-recons-19} to fit 3DMM to the video frames using a modified ResNet-50 network \cite{resnet}.
We retrain the network with masked face images as input, and the output is a combined vector $(\boldsymbol{\alpha}, \boldsymbol{\beta}, \boldsymbol{\delta}, \boldsymbol{\gamma}, \boldsymbol{p}) \in \mathbb{R}^{257}$, where $\boldsymbol{\alpha} \in \mathbb{R}^{80}$, $\boldsymbol{\beta} \in \mathbb{R}^{64}$, $\boldsymbol{\delta} \in \mathbb{R}^{80}$, $\boldsymbol{\gamma} \in \mathbb{R}^{27}$, and $\boldsymbol{p} \in \mathbb{R}^{6}$ represent face identity, expression, texture, illumination, and pose respectively. Concretely, the pose vector $\boldsymbol{p}$ is composed of a rotation vector\footnote{Euler angles \textit{yaw}, \textit{pitch}, and \textit{roll} for constructing the rotation matrix $\mathbf{R} \in \mathrm{SO}(3)$} $\boldsymbol{r} \in \mathbb{R}^{3}$ and a translation vector $\mathbf{t} \in \mathbb{R}^{3}$. 
With the predicted parameters, the shape $\mathbf{S}$ and texture $\mathbf{T}$ of the 3D face can be modeled as:
\begin{equation}\label{eq:3dmm-para}
\begin{split}
    \mathbf{S}&=\bar{\mathbf{S}}+\mathbf{B}_{id} \boldsymbol{\alpha}+\mathbf{B}_{exp} \boldsymbol{\beta}, \\
    \mathbf{T}&=\bar{\mathbf{T}}+\mathbf{B}_{tex} \boldsymbol{\delta},
\end{split}
\end{equation}
where $\bar{\mathbf{S}}$ and $\bar{\mathbf{T}}$ denote the mean shape and texture, $\mathbf{B}_{id}$, $\mathbf{B}_{exp}$, and $\mathbf{B}_{tex}$ are the PCA bases for face identity, expression, and texture respectively. Similar to Deng \etal~\cite{accu-face-recons-19}, we adopt $\bar{\mathbf{S}}$, $\bar{\mathbf{T}}$, $\mathbf{B}_{id}$, and $\mathbf{B}_{tex}$ from BFM~\cite{bfm}, and $\mathbf{B}_{exp}$ built from FaceWarehouse~\cite{faceware}.

\subsubsection{Texture Sampling and UV Mapping} 
Given the predicted shape $(\boldsymbol{\alpha}, \boldsymbol{\beta})$ and pose $(\boldsymbol{r}, \boldsymbol{t})$ parameters, we can transform a face from the image space to the UV space through texture sampling and UV mapping. We first project the 3D face model onto the image using the pose parameters and perform bilinear sampling to compute per-vertex texture for the 3D face model. For self-occluded and back-facing vertices, as well as vertices projected onto the masked regions, we simply assign zero to their texture values. Finally, we carry out UV mapping to transform the 3D face model texture to the UV space. 

For the rest of this paper, we denote a corrupted input frame and its ground truth as $\mathbf{I}_{in}$ and $\mathbf{I}_{gt}$ respectively, and their UV-maps as $\mathbf{U}_{in}$ and $\mathbf{U}_{gt}$ respectively. We represent the missing regions in the image space using a 2D binary mask $\mathbf{I}_{m}$, and denote its UV-map as $\mathbf{U}_{m}$. Similarly, we represent the valid projection of the 3D face model in the image space using a 2D binary mask $\mathbf{I}_{v}$, and denote its UV-map as $\mathbf{U}_{v}$ (see Fig.~\ref{fig-framework}). We also map the synthesized texture $\mathbf{T}$ to the UV space and denote it as $\mathbf{U}_{t}$.

\import{imgs/}{fig-uvattn.tex}

\subsection{Stage~I: UV-map Completion}\label{sec:Stage-I}
We first transform the face from the image space to the UV space and carry out UV-map completion. As mentioned previously, the UV maps of a face represent unwarped face textures which are well aligned and largely invariant to face poses and expressions. This greatly facilitates the learning of face structures and the finding of correspondences in neighboring frames.

\subsubsection{\uvmodel (\uvmodelAbbr)}
We adopt an encoder-decoder network equipped with gated convolutions \cite{gated-img} as the backbone of our \uvmodelAbbr (network details can be found in the supplementary material). We concatenate each frame $\mathbf{U}_{in}^{i}$ with its horizontally flipped UV-map $\hat{\mathbf{U}}_{in}^i$, synthesized texture map $\mathbf{U}_{t}^i$, valid face projection $\mathbf{U}_{v}^i$, and missing regions $\mathbf{U}_{m}^i$, and feed them to the encoder to generate the feature map $\mathbf{F}^i$:
\begin{equation}
    \mathbf{F}^i = En(\mathbf{U}_{in}^i, \hat{\mathbf{U}}_{in}^i, \mathbf{U}_{t}^i, \mathbf{U}_{v}^i, \mathbf{U}_{m}^i).
\end{equation}
The flipped UV map $\hat{\mathbf{U}}_{in}$ exploits symmetry to provide auxiliary information when only parts of the symmetrical face features are being masked, whereas the synthesized texture map $\mathbf{U}_{t}$ helps to provide auxiliary information when symmetrical face features are being completely masked. 

To exploit information provided by correspondences in neighboring frames, we propose a \uvattn (\uvattnAbbr) module to fuse features from neighboring frames. Specifically, for each {\em target frame}, we select $n$ other frames as its {\em reference frames} and fuse their features using the \uvattnAbbr module:
\begin{equation}
    \mathbf{Z}^i = Attn(\mathbf{F}^i, \{\mathbf{F}^{i+j}~|~j \in \Omega\}),
\end{equation}
where $\Omega$ is the set of offset indices for the reference frames.
In our experiments, we take $\Omega=\{-2,-1,+1,+2\}$.
Finally, the fused feature map $\mathbf{Z}^i$ is fed to the decoder to generate the completed UV map $\mathbf{U}_{out}^i$:
\begin{equation}
    \mathbf{U}_{out}^i = De(\mathbf{Z}^i).
\end{equation}

\subsubsection{\uvattn} 
Inspired by the recently proposed attention mechanism \cite{non-local}, we design a frame-wise attention block to explore correspondences between a target frame and its reference frames. Thanks to the well alignment of the face features in the UV space, we can limit our search for correspondences in a small local window. Concretely, we pick {\it query} points from the masked regions in the feature map of the target frame. For each {\it query}, we define a $s\times s$ small window (we set $s = 3$ in our experiments) centered at the {\it query} for selecting its {\it reference} points from the feature maps of the reference frames (see Fig.~\ref{fig-uvattn}). This small window design is employed to account for any slight misalignment of the UV maps. Given the query $\mathbf{q} \in \mathbb{R}^C$
evaluated at the query point, and the keys $\mathbf{K} \in \mathbb{R}^{C \times (s^2 \times n)}$
and values $\mathbf{V} \in \mathbb{R}^{C \times (s^2 \times n)}$ evaluated at the reference points, frame-wise attention is accomplished by
\begin{align}
    \begin{split}
        \bm{\alpha}&= \frac{\exp\left(\mathbf{K}^{\rm T} \mathbf{q}\right)}{\sum_{m=1}^{N}\exp\left(\mathbf{K}_{m}^{\rm T} \mathbf{q}\right)}, \\
        \mathbf{z}&= \mathbf{f} + W_{z}(\mathbf{V}\bm{\alpha}),
    \end{split}
\end{align}
where $N\!=\!s^2 \!\times\!n$ gives the total number of reference points; $\bm{\alpha}$, $\mathbf{f}$, $\mathbf{z}$, and $W_{z}$ denote the attention vector, input feature vector, output feature vector, and output embedding layers respectively. Compared with previous works such as STTN \cite{sttn} which uses spatial-temporal non-local attention to find correspondences across different frames, our design dramatically cuts down unnecessary computations and greatly improves the time complexity.

\subsubsection{Loss Functions}
We use $\mathcal{L}_{1}$ loss and SSIM loss \cite{ssim} for both UV maps and back-projected faces to train \uvmodelAbbr.
$\mathcal{L}_{1}$ loss aims at minimizing the distance between the ground-truth and predicted UV maps, whereas SSIM loss is adopted for maximizing structure similarity. 
The loss for the UV map is computed as
\begin{align}
\begin{split}
    \mathcal{L}_{U}&= \mathcal{L}^{U}_{1} + \mathcal{L}^{U}_{\mathit{SSIM}},\\
    \mathcal{L}^{U}_{1}&= \left\|\mathbf{U}_{v} \circ \left(\mathbf{U}_{out} - \mathbf{U}_{gt}\right) \right\|_{1} ,  \\
    \mathcal{L}^{U}_{\mathit{SSIM}}&= -
    \frac{\left(2 \mu_{\mathbf{U}_{\mathit{out}}} \mu_{\mathbf{U}_{\mathit{gt}}}+c_{1}\right)\!
    \left(2 \sigma_{\mathbf{U}_{\mathit{out}}} \sigma_{\mathbf{U}_{\mathit{gt}}}+c_{2}\right)}
    {\left(\mu_{\mathbf{U}_{\mathit{out}}}^{2} \! +\!\mu_{\mathbf{U}_{\mathit{gt}}}^{2} \! +\!c_{1}\right)\!
    \left(\sigma_{\mathbf{U}_{\mathit{out}}}^{2} \! +\!\sigma_{\mathbf{U}_{\mathit{gt}}}^{2} \! +\!c_{2}\right)} ,
\end{split}
\end{align}
where $\circ$ denotes element-wise multiplication, $\mu$ and $\sigma$ denote the mean and variance, $c_1$ and $c_2$ are stabilization constants set as $0.01^2$ and $0.03^2$ according to \cite{ssim}. 
Similarly, we define $\mathcal{L}^{I}_{1}$, $\mathcal{L}^{I}_{\mathit{SSIM}}$, and $\mathcal{L}_{I}$ for the back-projected face $\mathbf{I}_{bp}$ in the image space. The overall loss for Stage~I is given by
\begin{align}\label{eq:uv-recons}
    \begin{split}
        \mathcal{L}_{\rm (I)} 
        &= \lambda_{\alpha}  \cdot  \mathcal{L}_{U}
         + \lambda_{\beta}  \cdot  \mathcal{L}_{I},
    \end{split}
\end{align}
where the weights $\lambda_{\alpha}$ and $\lambda_{\beta}$ are empirically set as 1.0 and 2.0 respectively.

\import{tables/}{vd_compare.tex}

\input{imgs/fig-vd-sota-all}

\subsection{Stage~II: Face Video Refinement}\label{sec:Stage-II}
We transform the output from \uvmodelAbbr back to the image space by rendering the 3D face model with the predicted UV map, and denote this back-projected face as $\mathbf{I}_{bp}$. We then perform face video refinement to inpaint any background (non-face) regions not covered in Stage~I as well as to refine and fuse the inpainted face regions with the input frame.

\subsubsection{\refine (\refineAbbr)}
Similar to \uvmodelAbbr, we adopt an encoder-decoder network as the backbone for our \refineAbbr (networks details can be found in the supplementary material). We concatenate each frame $\mathbf{I}_{in}^i$ with its masked back-projected face $\mathbf{I}_{mbp}^i = \mathbf{I}_m^i \circ \mathbf{I}_{bp}^i$ and missing regions $\mathbf{I}_m^i$, and feed them to the encoder to generate the feature map $\tilde{\mathbf{F}}^i$. A Mask-wise Attention (MA) module is proposed to fuse features from non-masked regions in neighboring frames. MA block is similar to the FA block, but with the reference points taken from the non-masked regions of both the target and reference frames. The fused feature map is fed to the decoder to generate the predicted image $\mathbf{I}_{out}$. The final output $\mathbf{I}_{c}^i$ is then obtained by
\begin{align}
    \begin{split}
        \mathbf{I}_{c}^i = \mathbf{I}_{m}^i \circ \mathbf{I}_{out}^i + (1 - \mathbf{I}_{m}^i) \circ \mathbf{I}_{in}^i.
    \end{split}
\end{align}

\subsubsection{Loss Functions}
Similar to Stage~I, we adopt $\mathcal{L}^{I}_{\mathit{SSIM}}$ and a slightly modified version of $\mathcal{L}_{1}^{I}$ to train \refineAbbr. In addition, we also use perceptual loss $\mathcal{L}_{per}^{I}$ to minimize the distance in the semantic feature space. The overall loss for Stage~II is given by
\begin{align}
\begin{split}
    \mathcal{L}_{\rm (II)}&= \mathcal{L}_{1+}^{I}  
    + \mathcal{L}_{\mathit{SSIM}}^{I}
    + 0.1 \!\cdot\! \mathcal{L}_{per}^{I},
\end{split}
\end{align}
where
\begin{align}
\begin{split}
     \mathcal{L}^{I}_{1+}&= \left\|\mathbf{I}_{\mathit{out}}\! - \mathbf{I}_{\mathit{gt}} \right\|_{1}
        + 2\cdot \left\|\mathbf{I}_{m} \circ \left(\mathbf{I}_{\mathit{out}} \! - \mathbf{I}_{\mathit{gt}}\right) \right\|_{1}, \\
    \mathcal{L}^{I}_{\mathit{SSIM}}&= -
        \frac{\left(2 \mu_{\mathbf{I}_{\mathit{out}}} \mu_{\mathbf{I}_{\mathit{gt}}}+c_{1}\right)\!
        \left(2 \sigma_{\mathbf{I}_{\mathit{out}}} \sigma_{\mathbf{I}_{\mathit{gt}}}+c_{2}\right)}
        {\left(\mu_{\mathbf{I}_{\mathit{out}}}^{2} \! +\!\mu_{\mathbf{I}_{\mathit{gt}}}^{2} \! +\!c_{1}\right)\!
        \left(\sigma_{\mathbf{I}_{\mathit{out}}}^{2} \! +\!\sigma_{\mathbf{I}_{\mathit{gt}}}^{2} \! +\!c_{2}\right)},\\
    \mathcal{L}^{I}_{per}&= \frac{1}{C_k H_k W_k}
        \left\|\phi_{k}\left(\mathbf{I}_{\mathit{out}}\right) - \phi_{k}\left(\mathbf{I}_{\mathit{gt}}\right)\right\|_{2}^{2},
\end{split}
\end{align}
where $\phi_{k}$ is the $k$-th layer output of a pretrained VGG-16 network \cite{vgg}, $C_k$, $H_k$, and $W_k$ denote the channel number, height, and width of the $k$-th layer output respectively.

%% file: imgs/fig-uvattn.tex
\begin{figure}[t]
\begin{center}
   \includegraphics[width=\linewidth]{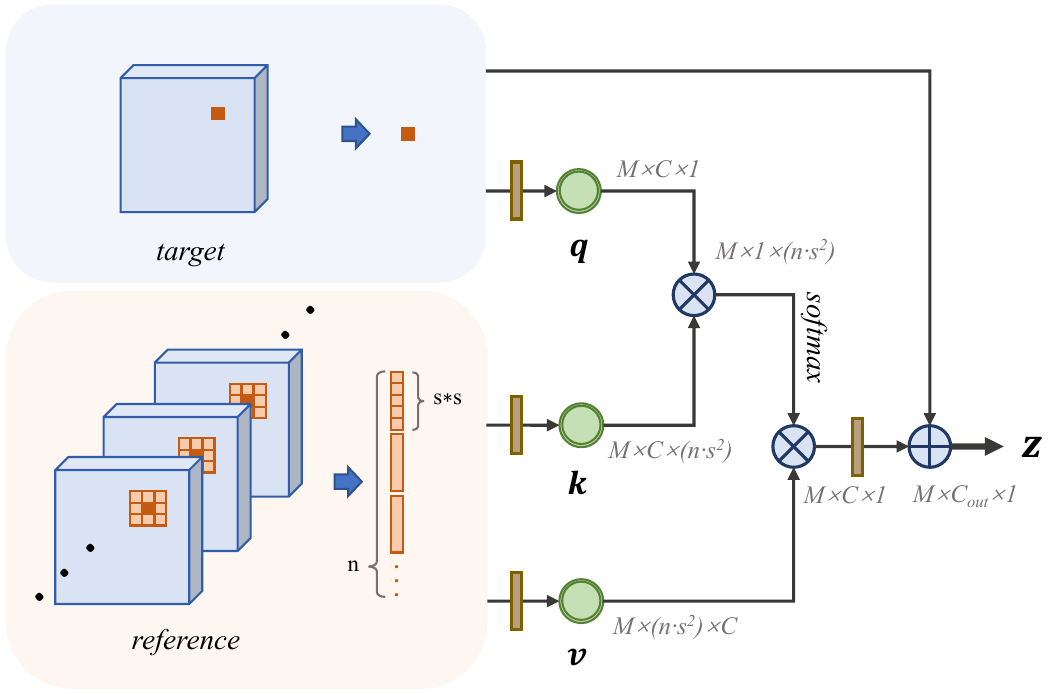}
\end{center}
   \caption{Detailed structure of our \uvattn~(\uvattnAbbr) module. The symbol ``$\otimes$'', ``$\oplus$'' in the figure represent multiplication and element-wise sum, respectively.
    $M$ is the number of points in the masked regions (embedding space); 
$C$ is the channel dimension; 
$n \cdot s^2$ gives the number of selected memory points (key and value) from the reference frames. 
   }
\label{fig-uvattn}
\end{figure}

%% file: tables/vd_compare.tex
\begin{table*}
\caption{Comparison with the state-of-the-art video inpainting methods.
}
\label{table:vd-sota}
\begin{center}
\setlength{\tabcolsep}{3pt}
\resizebox{\textwidth}{!}{
\begin{tabular}{l||
*{4}{@{}p{1cm}<{\centering}@{}}|*{4}{@{}p{1cm}<{\centering}@{}} ||
*{4}{@{}p{1cm}<{\centering}@{}}|*{4}{@{}p{1cm}<{\centering}@{}}
}
\noalign{\hrule height 1pt}
\multirow{3}{1cm}{Method}
 & \multicolumn{8}{c||}{Rectangular Mask}
 & \multicolumn{8}{c}{Irregular Mask} 
 \TBstrut\\
\cline{2-17}
 & \multicolumn{4}{c|}{Shifting Mask} 
 & \multicolumn{4}{c||}{Static Mask} 
 & \multicolumn{4}{c|}{Shifting Mask} 
 & \multicolumn{4}{c}{Static Mask} 
 \TBstrut\\
 &  $\ell_{1}\downarrow$ & PSNR$\uparrow$  & SSIM$\uparrow$  &VFID$\downarrow$    
 &  $\ell_{1}\downarrow$ & PSNR$\uparrow$  & SSIM$\uparrow$  &VFID$\downarrow$ 
 &  $\ell_{1}\downarrow$ & PSNR$\uparrow$  & SSIM$\uparrow$  &VFID$\downarrow$    
 &  $\ell_{1}\downarrow$ & PSNR$\uparrow$  & SSIM$\uparrow$  &VFID$\downarrow$ 
       \Bstrut\\
 \hline 
DeepfillV2~\cite{gated-img} 
        & 0.0624 & 21.22 & 0.9470  &  0.2826
        & 0.0665 & 20.66 & 0.9569  &  0.2751
        & 0.0665 & 20.74 & 0.9358  &  0.5649
        & 0.0567 & 22.43 & 0.9694  &  0.3847
        \Tstrut\\
LaFin~\cite{lafin-img-face} 
        & 0.0658 & 21.19 & 0.9450  &  0.2975
        & 0.0724 & 20.41 & 0.9529  &  0.3283
        & 0.0647 & 21.35 & 0.9400  &  0.3701
        & 0.0597 & 22.37 & 0.9693  &  0.3036
        \\
STN-GAN~\cite{img2vd-vd-face} 
        & 0.0644 & 21.23 & 0.9505  &  0.2547
        & 0.0686 & 20.57 & 0.9573  &  0.3153
        & 0.0574 & 21.90 & 0.9494  &  0.3120
        & 0.0524 & 23.14 & 0.9745  &  0.2330
        \\
Co-Mod-GAN~\cite{zhao2021large}
 & 0.0784 & 20.16 & 0.9382 &    0.4341
 & 0.0780 & 20.11 & 0.9519 &    0.4003
 & 0.0702 & 20.85 & 0.9398 &    0.4298
 & 0.0634 & 22.07 & 0.9691 &    0.3358
  \\
MAT \cite{li2022mat}
 & 0.0782 & 19.75 & 0.9345 &    0.4046
 & 0.0937 & 18.24 & 0.9402 &    0.6562
 & 0.0739 & 20.06 & 0.9309 &    0.3960
 & 0.0636 & 21.46 & 0.9660 &    0.2840
  \\
VINet~\cite{deep-vd-inpaint} 
        & 0.0779 & 21.72 & 0.9556  &  0.1772
        & 0.1354 & 17.34 & 0.9402  &  0.5028
        & 0.0887 & 20.30 & 0.9461  &  0.2246
        & 0.1463 & 18.03 & 0.9516  &  0.3749
        \\
3DGated~\cite{3dconv-vd} 
        & 0.0436 & 24.40 & 0.9658  &  0.1874
        & 0.0663 & 20.82 & 0.9574  &  0.4170
        & 0.0437 & 24.40 & 0.9598  &  0.2757
        & 0.0510 & 23.48 & 0.9729  &  0.2699
        \\
STTN~\cite{sttn}  
        & 0.0385 & 25.18 & 0.9684  &  0.1829
        & 0.0571 & 21.89 & 0.9627  &  0.3372
        & 0.0389 & 24.94 & 0.9655  &  0.2235
        & 0.0484 & 23.49 & 0.9756  &  0.2213
        \\
FuseFormer~\cite{liu2021fuseformer}
 & 0.0508 & 22.98 & 0.9558 &   0.2745
 & 0.0537 & 22.52 & 0.9647 &   0.2685
 & 0.0491 & 23.03 & 0.9529 &   0.3265
 & 0.0424 & 24.71 & 0.9778 &   0.2102
  \\
Ours  
        & \textbf{0.0372} & \textbf{25.36} & \textbf{0.9709}  &  \textbf{0.1587}
        & \textbf{0.0466} & \textbf{23.64} & \textbf{0.9710}  &  \textbf{0.2649}
        & \textbf{0.0342} & \textbf{25.92} & \textbf{0.9706}  &  \textbf{0.1813}
        & \textbf{0.0344} & \textbf{26.41} & \textbf{0.9832}  &  \textbf{0.1816}
        \Bstrut\\
\noalign{\hrule height 1pt}
\end{tabular}
} 
\end{center}
\end{table*}

%% file: imgs/fig-vd-sota-all.tex
\begin{figure*} \centering
    \includegraphics[align=c,width=\textwidth]{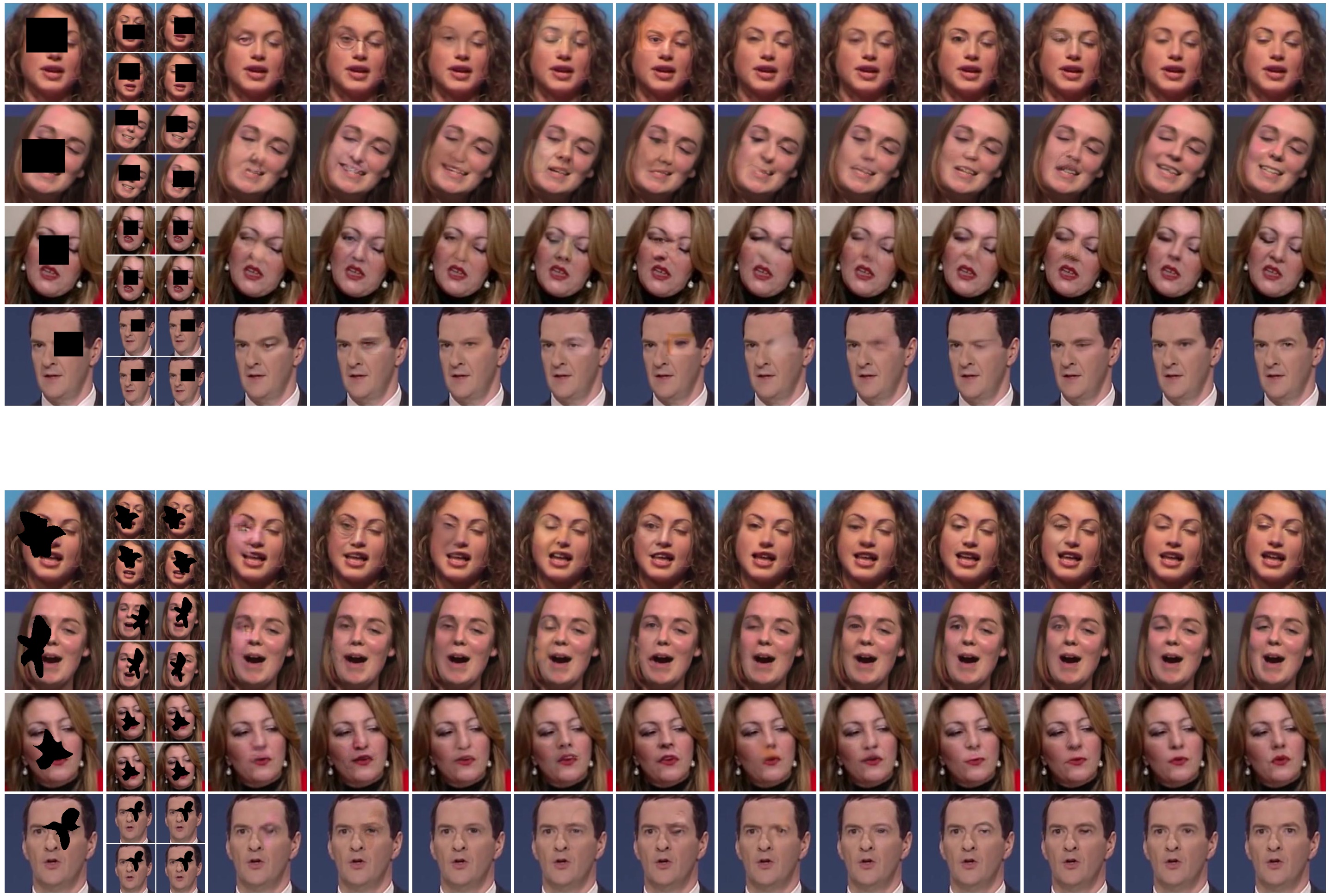}\\
    \vspace{-0.691\textwidth}
    {\small {\bf Rectangular Mask} (from top to bottom: case A, B, C, D)}\\
    \vspace{0.29\textwidth}
    \subfloat{\resizebox{\linewidth}{!}{
    \begin{tabular}{@{}*{13}{>{\centering\arraybackslash}p{2.1cm}}}
Input & Reference & DeepfillV2 & LaFin & STN-GAN & Co-Mod-GAN & MAT & VINet & 3DGated & STTN & FuseFormer & Ours & Ground Truth
    \end{tabular}}}
    \vspace{0.02\textwidth}
    {\small {\bf Irregular Mask} (from top to bottom: case A, B, C, D)}\\
    \vspace{0.29\textwidth}
    \subfloat{\resizebox{\linewidth}{!}{
    \begin{tabular}{@{}*{13}{>{\centering\arraybackslash}p{2.1cm}}}
Input & Reference & DeepfillV2 & LaFin & STN-GAN & Co-Mod-GAN & MAT & VINet & 3DGated & STTN & FuseFormer & Ours & Ground Truth
    \end{tabular}}} 
\caption{Comparison with the state-of-the-arts. The top four columns are rectangular mask cases and bottom four are irregular mask cases. 
}
\label{fig-vd-sota-all}
\end{figure*}

%% file: text/4_experiments.tex

\import{tables/}{vd_ablation.tex}

\section{Experiments}

\subsection{Implementation Details}

\subsubsection{Dataset}
We use the 300VW~\cite{300vw} dataset for our experiments. 
300VW dataset contains 114 face videos with diverse face poses and expressions. We excluded low quality videos and selected 75 videos for training and 20 for evaluation. 

\subsubsection{Inpainting Settings}
We followed the pre-processing described in Deng \etal~\cite{accu-face-recons-19} to crop and resize the face regions.
The image size adopted for face videos is $224\times 224$ and the UV maps have a dimension of $256\times 256$. 
To verify our contribution in handling large pose variations, we extracted every $10$-th frames from the original face videos as our test sequences. 
Since our method is robust to pose variations, we randomly sample reference frames from the whole sequence for the training phase.

\subsubsection{Mask Settings}
We consider two kinds of masks for evaluation:
\begin{itemize}
\renewcommand\labelitemi{--}
    \item \textbf{Shifting masks} are generated with slightly altered shapes and quick motions across frames, which mimic non-stationary occlusions in face videos.
    \item \textbf{Static masks} keep consistent shapes and locations for the whole video sequence, which also commonly happen in real scenes.
\end{itemize}

We also consider two kinds of mask shapes:
\begin{itemize}
\renewcommand\labelitemi{--}
    \item \textbf{Rectangular masks} are a representative case which is commonly used in inpainting tasks.
    \item \textbf{Irregular masks}~\cite{sttn} mimic arbitrarily shaped occlusion objects in face videos. 
\end{itemize}
The generated masks occupy between $8\%$-$20\%$ of the whole image. Irregular masks are only evaluated in the baseline comparison (Sec.~\ref{sec:sota}). We test both mask shapes under the shifting and static cases.

\subsubsection{Metric Settings}
We consider four different metrics in our quantitative evaluations, namely 
(1) $\ell_{1}$ error; 
(2) PSNR (Peak Signal-to-Noise Ratio); 
(3) SSIM~\cite{ssim} (Structural Similarity); and 
(4) VFID~\cite{3dconv-vd,wang2018vid2vid} (Video-based Fr\'echet Inception Distance, a video perceptual measure).

\import{imgs/}{fig-vd-ablation.tex}

\import{imgs/}{fig-uv-ablation.tex}

\import{tables/}{uv_ablation.tex}

\import{imgs/}{fig-uvattn-viz.tex}

\subsection{Comparison with State-of-the-Arts}\label{sec:sota}

In this section, we conducted comparison between the proposed method and other inpainting methods to illustrate the strength of our framework.

\subsubsection{Baselines}
To the best of our knowledge, limited works have been proposed for face videos and consider the combination of face prior with video inpainting pipeline.
We therefore look for video inpainting works that have been tested on face videos with codes publicly available, and select \cite{3dconv-vd} for comparison.
Apart from~\cite{3dconv-vd}, we also select two representative video inpainting baselines~\cite{deep-vd-inpaint,sttn} and two image inpainting works~\cite{lafin-img-face,gated-img} for comparison. 
To evaluate the effectiveness of our method on face videos, we also reimplement a face video re-identification method~\cite{img2vd-vd-face} for comparison.
All the baselines are recent deep learning methods developed for general scenes or face images. Below gives a brief summary of them:
\begin{itemize}
\renewcommand\labelitemi{--}
    \item \textbf{DeepfillV2}~\cite{gated-img}, an encoder-decoder structured method based on 2D gated convolutions and contextual attention.
    \item \textbf{LaFin}~\cite{lafin-img-face}, a landmark-guided two-stage method proposed for face image inpainting. 
    \item \textbf{STN-GAN}~\cite{img2vd-vd-face}, a GAN-based model proposed for face re-identification using 3D Residual blocks to aggregate features. 
    \item \textbf{Co-Mod-GAN}~\cite{zhao2021large}, a GAN-based image inpainting method with co-modulation of both conditional and stochastic style representations.
    \item \textbf{MAT}~\cite{li2022mat}, a transformer-based model for large hole image inpainting.
    \item \textbf{VINet}~\cite{deep-vd-inpaint}, a context aggregation method based on recurrent structures and flow-warping. 
    \item \textbf{3DGated}~\cite{3dconv-vd}, an encoder-decoder network based on 3D gated convolutions.
    \item \textbf{STTN}~\cite{sttn}, a transformer-based method by spatial and temporal patch-matching.
    \item \textbf{FuseFormer}~\cite{liu2021fuseformer}, a transformer-based method with fine-grained feature fusion for video inpainting.
\end{itemize}
For a fair comparison, we retrained their models on the same dataset using their publicly-available code. Since codes are not available for STN-GAN~\cite{img2vd-vd-face}, we reimplemented their method according to the details in their paper. However, due to the nature of their task, they require ground truth landmarks as additional inputs, which are not available for face inpainting tasks. We therefore trained a landmark prediction network\cite{bulat2017far} to predict landmarks from the corrupted faces for them.
Since the model design of Co-Mod-GAN \cite{zhao2021large} and MAT \cite{li2022mat} requires the resolution of input images to be a power of 2, We resize the input images to 256. Therefore, quantitative results may be affected to some extend.

\subsubsection{Quantitative Comparison}
Table~\ref{table:vd-sota} summarizes the quantitative comparison results, where our method consistently outperformed the other methods on all four metrics under the two different mask settings. Due to the lack of face priors, all three video baselines fail to reconstruct the faces under the static mask setting. Due to the difficulty of correspondence retrieval in face videos with large pose / expression variations, they also performed poorly in the shifting mask setting. 
For image-based methods, even though face prior was utilized, they still failed since no temporal information was considered. Further, it is observed that the performance of landmark guided method may be affected by the limited accuracy of the landmarks predicted from corrupted faces.

\subsubsection{Qualitative Comparison}
We further conducted visual comparison on four classic scenes:
\begin{itemize}
    \item[(A)] Face expression appears differently in reference frames; 
    \item[(B)] Face pose changes frequently; 
    \item[(C)] No useful reference in other frames (\eg, static masks);
    \item[(D)] No useful reference in other frames, however, it can be self-referenced (\eg, one eye covered).
\end{itemize}
Results are shown in Fig.~\ref{fig-vd-sota-all} for both rectangular masks and irregular masks. For each kind of masks, case A, B, C, and D are presented from top to bottom.

Since we target at face videos, where correspondence retrieval is much more difficult than general scenes due to large face pose and expression variations, all the video baselines failed in these challenging cases (A \& B).
Specifically, in case A, other video-based methods either attended to or directly copied the opened eyes from the reference frames and produced incorrect results. In case B, when face pose varied largely between frames, even though reference could be retrieved from other frames, they failed to comprehend the 3D face structure and directly incorporated the nose under a different pose to the target frame, resulting in a distorted face.

For case C and case D, due to the lack of face prior, they all failed to predict proper face structures when no useful reference could be obtained (though it could be self-referenced in case D). The flow-based context aggregation method VINet failed completely in the static mask setting. As expected, our method performed the best on these challenging cases and achieved the most visually pleasant results compared to the other baselines. Through the use of 3D face prior, our method can take full advantage of the well alignment and symmetry properties of the UV maps and robustly restore the missing face regions even under large face pose and expression variations.

For methods targeting at single face~\cite{3dconv-vd,lafin-img-face}, they do not retrieve useful information from other frames but merely synthesize the missing regions for the current frame. Hence, they treat all the testing cases (shifting \& static) the same way. It is obvious that they all failed to generate temporal consistent contents for the missing regions. 
Note that LaFin~\cite{lafin-img-face} and STN-GAN~\cite{img2vd-vd-face} also utilize face prior (i.e., landmarks) as their guidance. However, since the inpainting branch heavily depends on the landmark detection results, it will generate obvious artifacts when the predicted landmarks are incorrect (see Fig.~\ref{fig-vd-sota-all}).

\subsection{Analysis of the Proposed Framework}

In this section, we present experimental results to verify the design of our framework.

\subsubsection{Effectiveness of UV-map Completion}
We first carried out analysis on the effectiveness of our UV-map completion stage. We considered three variants, namely (a) single frame without UV maps as guidance, (b) single frame with UV maps as guidance, and (c) multi-frame without UV maps as guidance. Results are shown in Fig.~\ref{fig-vd-ablation} and Table~\ref{table:vd-ablation}. Our full model achieved the most plausible results compared to these variant models. 
It is also observed that the performance improved considerably with UV maps as guidance especially under the static mask setting.

\subsubsection{Effectiveness of FA Module}
We also conducted ablation study to evaluate our FA module. For comparison, we considered three different baselines, namely (a) simply taking a single frame as input, (b) concatenating the target frame with its reference frames as input, and (c) fusing (concatenating) the features of all the frames in the latent space before decoder. The quantitative analysis evaluated on $\mathbf{U}_{out}$ are listed in Table~\ref{table:uv-ablation}. Our method achieved the best performance with the assistance of \uvattn. Referring to the qualitative results shown in Fig.~\ref{fig-uv-ablation}, it is observed that our full model outperformed all the others in both detail generation and texture consistency, which also demonstrates the effectiveness of the FA module in retrieving proper correspondences for corrupted regions.

\subsubsection{Visualization of \uvattn}
To further investigate how does the \uvattnAbbr module work, we present the visualization of the \uvattn in Fig.~\ref{fig-uvattn-viz}. 
We labeled each reference frame with a distinct color to visualize the attention map in a more intuitive way. 
For each {\it query} point from the embedded features in the target frame, we selected the most responsive {\it key} point (maximum attention value) from its pool of {\it key} candidates, and filled the attention map with the index color of the corresponding reference frame.
In this example, we used the colors \{{\it \textcolor{Red}{red}}, {\it \textcolor{Green}{green}}, {\it \textcolor{Blue}{blue}}, {\it \textcolor{Dandelion}{yellow}}\} to denote the reference frames from left to right. 
The attention distribution is shown in the first column with the representative colors, while in the right four columns we display the response map of each reference frame.
From the attention distribution, it is observed that the model learns to retrieve the matching features from the regions with higher reliability, \ie, intact regions. 
With the \uvattnAbbr module, our proposed \uvmodelAbbr can better exploit the reference features and generate more visually plausible content for the corrupted face.

\import{tables}{patch-size.tex}

\import{imgs/}{fig-patch-mean.tex}

\import{imgs/}{fig-patch-size.tex}

\subsubsection{Ablation Study on UV-map completion Stage}
As mentioned in Sec.~\ref{sec:Stage-I}, we take both flipped UV map $\Hat{\mathbf{U}}_{in}$, the synthesized texture map $\mathbf{U}_{t}$, and the valid projection $\mathbf{U}_{v}$ as input. To further evaluate their contributions, we conducted ablation study on these components. Both quantitative results in Table~\ref{table:uv-ablation} and qualitative results in Fig.~\ref{fig-uv-ablation} demonstrate their effectiveness in reconstructing the face textures by utilizing the symmetry prior ($\Hat{\mathbf{U}}_{in}$) and 3D face model prior ($\mathbf{U}_{t}$).
While $\mathbf{U}_{v}$ which indicates the valid face regions of the UV texture can help stabilize the training process and improve the overall performance.

\subsubsection{Analysis on Patch Size used in \uvattn Module}
Our method utilizes 3DMM face model as a bridge to transform the face textures from image space to UV space.
Though the retrained face reconstruction network is capable of reconstructing proper face shapes for the corrupted input faces (refer to supplementary), it is possible that the predicted faces are slightly misaligned, which may result in small misalignment in the transformed UV maps.
Fig.~\ref{fig-patch-mean} shows the mean value of a bunch of inputs (target frame and its reference frames). We can see that there exists some small inconsistency especially around the eye regions.
Therefore, for each \textit{query} pixel, we propose to extract reference features in a local $s\times s$ window across all the reference frames. 
We further analyzed the effects of different patch sizes on shifting masks to observe how it affects the correspondence retrieval efficiency. Qualitative and quantitative results are shown in Fig.~\ref{fig-patch-size} and Table~\ref{table:patch-size} respectively.  
It is observed that adopting local windows instead of a single point can benefit the correspondence retrieval (the attention is more concentrated instead of scattered across the frames) and improve the overall performance. In our experiments, we adopted patch size $s=3$ to achieve a balance between performance and efficiency.

\subsubsection{Speed}
We also estimated the processing speed of our method to assess its applicability.
Our model achieved 19.3 fps with an NVIDIA GTX 2080Ti GPU card.
Despite the primary goal of improving the inpainting quality for face videos, our method still achieves reasonable efficiency  with a na\"ive implementation.
Specifically, Resnet-50 as feature extractor occupies $3.3\%$ of the time consumption, and the two main networks \uvmodelAbbr and \refineAbbr take $45.4\%$ in total, 
while the remaining $51.3\%$ are for rendering process in UV mapping.
We also compare our method with three recent video inpainting methods on inference speed. Results are shown in Table \ref{table:vd-speed}. 
Please note that the rendering process in UV mapping accounts for $51.3\%$ of the total time consumption. Nevertheless, our inference speed is still comparable to other baselines.  In real applications, the rendering process can be further optimized and greatly accelerated.

\input{tables/vd_speed}

\import{imgs}{fig-user-study.tex}

\subsection{User Study}
We conducted a user study to further evaluate the visual quality of the inpainted videos. For comparison, we chose one image-based method LaFin~\cite{lafin-img-face} with landmarks as guidance, and two video-based methods -- 3DGated~\cite{3dconv-vd}, and STTN~\cite{sttn} with relatively higher performance. 
We sampled 16 videos from the test dataset, and tested on both static mask and shifting mask to evaluate the performance on these two cases. For each case, we sampled clips lasting 10 seconds from either rectangle mask or irregular mask (8 for each).
The comparison is conducted in one-to-one manner with totally $3\times2\times16=96$ questions.
For each question, the volunteers were given both the masked video and ground-truth video for reference, and were required to pick the better one from two inpainted videos (one baseline and ours).
We collected responses from 20 volunteers and visualized the results in percentage (see Fig.~\ref{fig-user-study}). 
Our method gained most of the preference compared to other methods, which further demonstrates the effectiveness of our method.

\subsection{Application}
Face video inpainting usually serves as a recovering tool in many applications, such as video editing or restoration. It can be used to remove unwanted watermark / subtitles or objects that appear in face videos. An example is shown in Fig.~\ref{fig-application} demonstrating the watermark removal application. Since our method is capable of both shifting and static masks with arbitrary shapes, it can benefit diverse face video editing tasks especially for those with large pose / expression variations (e.g., talk show).

\import{imgs}{fig-application.tex}
\import{imgs}{fig-failure.tex}

\subsection{Failure Case \& Future Work}
Since our method utilizes face model to explore the underlying 3D structure of the given corrupted faces, it is possible the predicted 3DMM is not perfectly fitted to the ground truth face, especially when the mask covers key clues for accurate alignment. As shown in Fig.~\ref{fig-failure}, the eyes and nose are masked in the profile face, thus making it ambiguous for face reconstruction. The misaligned 3DMM (especially for nose region) results in noisy texture in the UV map and distorted nose in the final output. 
Currently, our second stage can help deal with small misalignment to refine the results. In our future work, we will try to improve the robustness of masked face reconstruction. Moreover, we will also extend this work to high-resolution face videos.

%% file: tables/vd_ablation.tex
\begin{table*}
\caption{Quantitative analysis on effectiveness of UV-map completion.
Here ``singles'' means single frame and ``multi'' means multi-frame.
}
\label{table:vd-ablation}
\begin{center}
\begin{tabular}{l||
*{4}{c}|*{4}{c}}
\hline
\multirow{2}{4.5em}{Method}
 & \multicolumn{4}{c|}{Shifting Mask} 
 & \multicolumn{4}{c}{Static Mask} \TBstrut\\

 &  $\ell_{1}\downarrow$ & PSNR $\uparrow$  & SSIM $\uparrow$  &VFID $\downarrow$    
 &  $\ell_{1}\downarrow$ & PSNR $\uparrow$  & SSIM $\uparrow$  &VFID $\downarrow$ 
       \Bstrut\\
 \hline\hline
single, w/o $\mathbf{I}_{mbp}$  \ 
        & 0.0544 & 22.53 & 0.9555  & 0.2547
        & 0.0586 & 21.94 & 0.9627  & 0.3293
        \Tstrut\\
single \ 
        & 0.0513 & 23.12 & 0.9586  & 0.2425
        & 0.0541 & 22.78 & 0.9660  & 0.2889
        \\
multi, w/o $\mathbf{I}_{mbp}$     
        & 0.0395 & 24.89 & 0.9676  & 0.1848
        & 0.0559 & 22.04 & 0.9638  & 0.3917
        \\
Ours    
        & \textbf{0.0372} & \textbf{25.36} & \textbf{0.9709} & \textbf{0.1587}
        & \textbf{0.0466} & \textbf{23.64} & \textbf{0.9710} & \textbf{0.2649}
        \Bstrut\\
\hline
\end{tabular}
\end{center}
\end{table*}

%% file: imgs/fig-vd-ablation.tex
\begin{figure}[t]
\begin{center}
  \includegraphics[width=\linewidth]{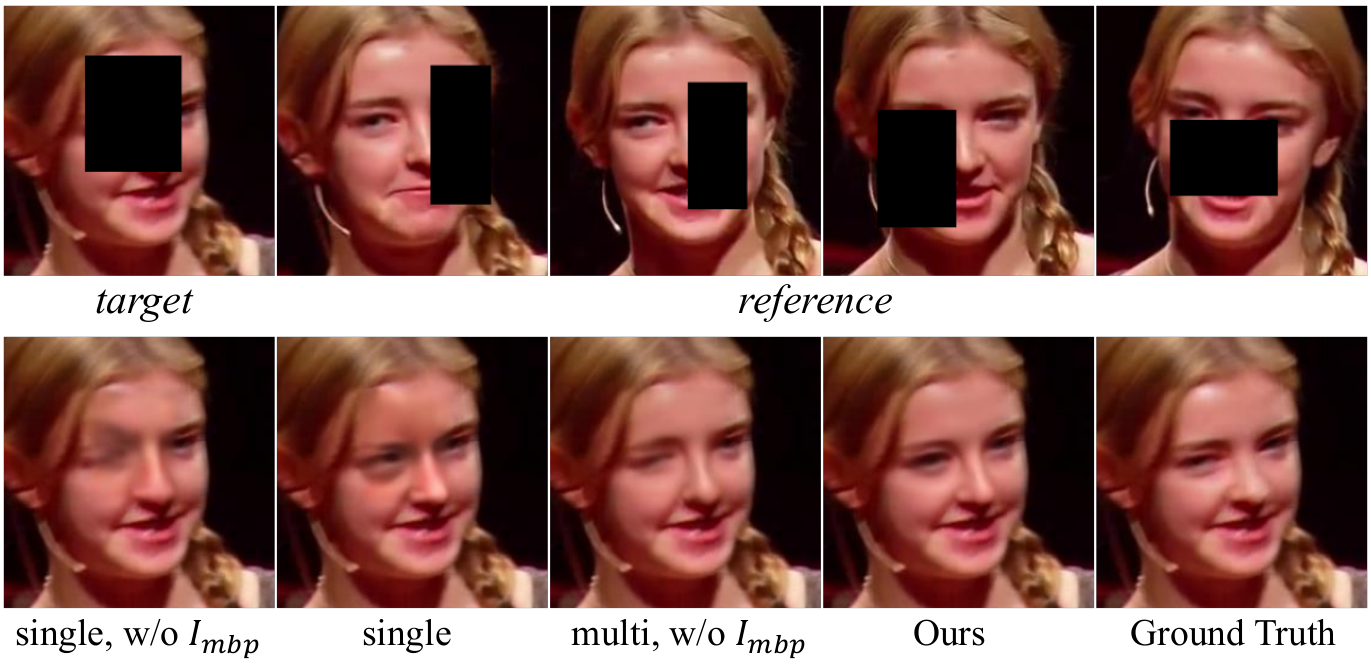}
\end{center}
\caption{Qualitative analysis on effectiveness of UV-map completion. 
}
\label{fig-vd-ablation}
\end{figure}

%% file: imgs/fig-uv-ablation.tex
\begin{figure*}
\begin{center}
   \includegraphics[width=0.98\linewidth]{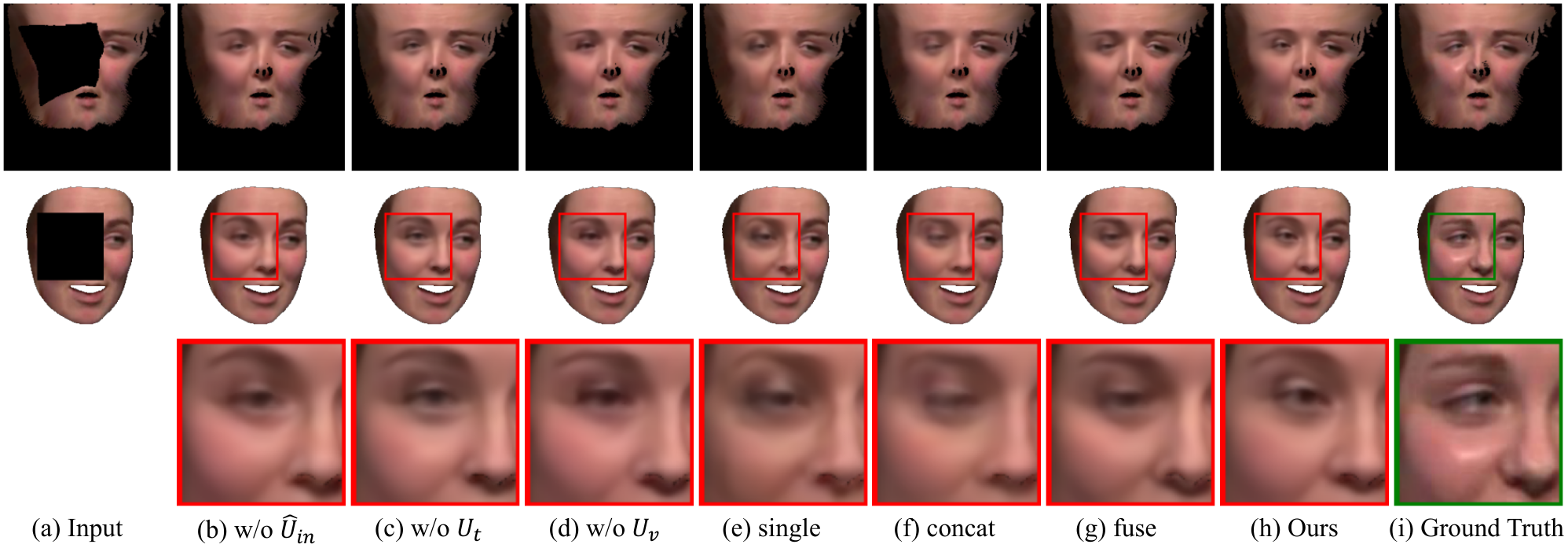}
\end{center}
   \caption{Qualitative results of the ablation study on UV-map completion. Here from top to bottom: UV map, back-projected face, and the highlighted region.}
\label{fig-uv-ablation}
\end{figure*}

%% file: tables/uv_ablation.tex
\begin{table*}
\caption{Analysis of UV-map completion stage. 
Here ``single'' means trained with single frame as input; ``concat'' means using concatenated frames as input; ``fuse'' means fusing (concatenating) the feature maps of all $n+1$ frames before the decoder.
}
\label{table:uv-ablation}
\begin{center}
\begin{tabular}{l||
*{4}{c}|*{4}{c}}
\hline
\multirow{2}{4em}{Method}
 & \multicolumn{4}{c|}{Shifting Mask} 
 & \multicolumn{4}{c}{Static Mask} \TBstrut\\

 &  $\ell_{1}\downarrow$ & PSNR $\uparrow$  & SSIM $\uparrow$  &VFID $\downarrow$    
 &  $\ell_{1}\downarrow$ & PSNR $\uparrow$  & SSIM $\uparrow$  &VFID $\downarrow$ 
       \Bstrut\\
 \hline\hline
 w/o $\Hat{\mathbf{U}}_{in}$ 
        & 0.0399 & 24.68 & 0.9723  &  0.2042
        & 0.0495 & 23.27 & 0.9726  &  0.2976
        \Tstrut\\
 w/o $\mathbf{U}_{t}$ 
        & 0.0397 & 24.70 & 0.9728  &  \textbf{0.1981}
        & 0.0505 & 23.05 & 0.9725  &  0.2887
        \\
 w/o $\mathbf{U}_{v}$       
        & 0.0400 & 24.72 & 0.9726  &  0.2058
        & 0.0496 & 23.33 & 0.9729  &  0.2901
        \Bstrut\\
\hline
single      
        & 0.0520 & 22.93 & 0.9639  &  0.2724
        & 0.0542 & 22.73 & 0.9699  &  0.3095
        \Tstrut\\
concat  
        & 0.0459 & 23.84 & 0.9679  &  0.2416
        & 0.0538 & 22.79 & 0.9701  &  0.3121
        \\
fuse   
        & 0.0420 & 24.36 & 0.9708  &  0.2077
        & 0.0509 & 23.11 & 0.9716  &  0.2991
        \Bstrut\\
\hline
Ours    
        & \textbf{0.0394} & \textbf{24.80} & \textbf{0.9731}  &  0.2007
        & \textbf{0.0483} & \textbf{23.47} & \textbf{0.9735}  &  \textbf{0.2824}
        \TBstrut\\
\hline
\end{tabular}
\end{center}
\end{table*}

%% file: imgs/fig-uvattn-viz.tex
\begin{figure}[t]
\begin{center}
   \includegraphics[width=0.99\linewidth]{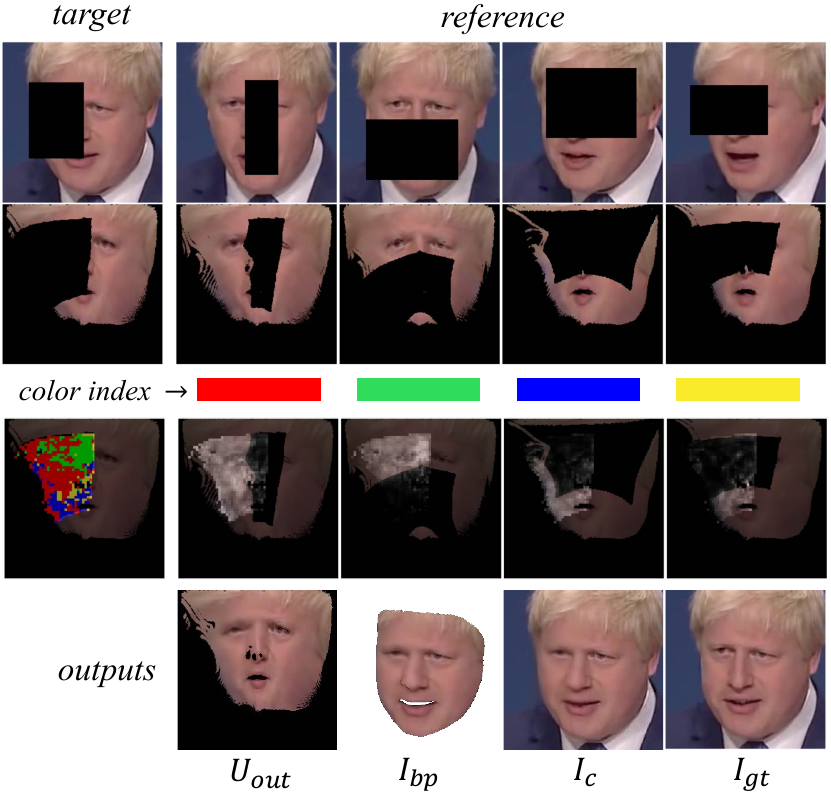}
\end{center}
  \caption{Visualization of our \uvattn\ module. From top to bottom: input frame $\mathbf{I}_{in}$, input UV map $\mathbf{U}_{in}$, color index, attention map,
  and outputs. For the upper part, the first column is the \textit{target} frame, the right four columns are \textit{reference} frames. 
}
\label{fig-uvattn-viz}
\end{figure}

%% file: tables/patch-size.tex
\begin{table}[t]
\caption{Quantitative results of different patch sizes on shifting masks.
}
\label{table:patch-size}
\begin{center}
\begin{tabular}{l||
*{4}{c}
}
\hline
Patch Size
 &  $\ell_{1}\downarrow$ & PSNR $\uparrow$  & SSIM $\uparrow$  &VFID $\downarrow$    
       \TBstrut\\
 \hline\hline
$s=1$      
        & 0.0404 & 24.63 & 0.9720  &  0.2161
        \Tstrut\\
$s=3 \text{(ours)}$ 
        & \textbf{0.0394} & \textbf{24.80} & \textbf{0.9731}  &  \textbf{0.2007}
        \\
$s=5$  
        & 0.0398 & 24.76 & 0.9727 &  0.2053
        \\
$s=7$   
        & 0.0398 & 24.69 & 0.9727  &  0.2104
        \Bstrut\\
\hline
\end{tabular}
\end{center}
\end{table}

%% file: imgs/fig-patch-mean.tex
\begin{figure}[t]
\begin{center}
  \includegraphics[width=.99\linewidth]{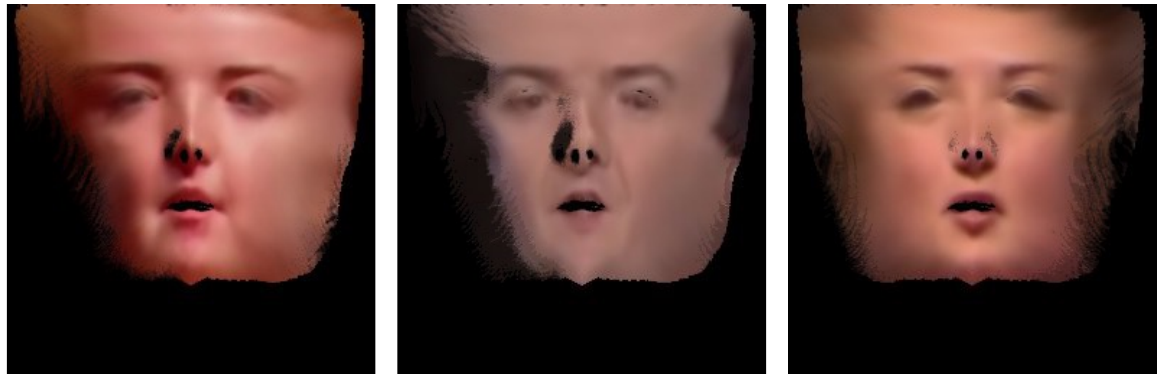}
\end{center}
\caption{Three examples of the averaged UV maps (per-pixel mean values of 5 consecutive frames).  Small misalignment can be observed  around  the  eye  regions.
}
\label{fig-patch-mean}
\end{figure}

%% file: imgs/fig-patch-size.tex
\begin{figure*}
\begin{center}
  \includegraphics[width=.95\linewidth]{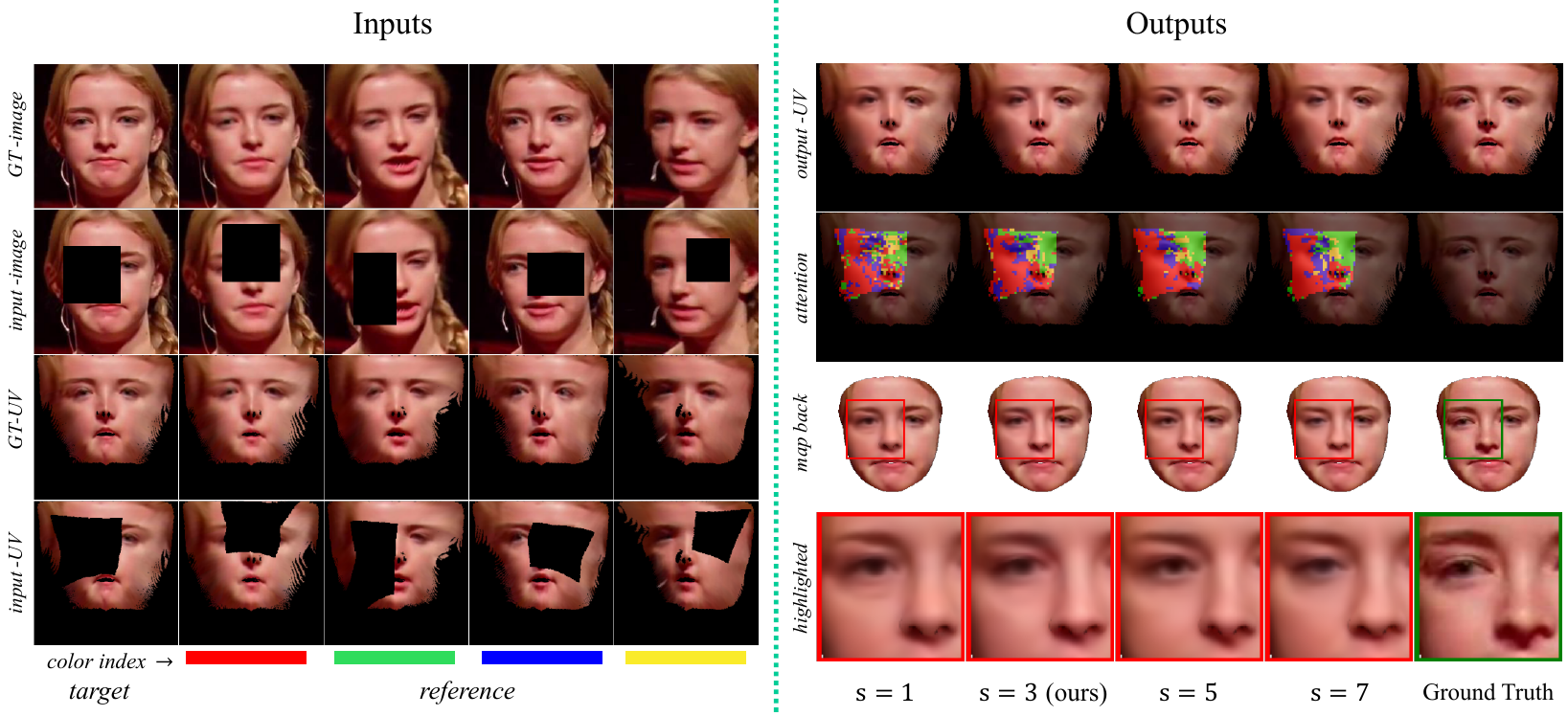}
\end{center}
\caption{Qualitative analysis on different patch size used in our \uvattn module. The left columns shows the masked input and ground truth of the {\it target} frame and {\it reference} frames, where from top to bottom: ground-truth image, masked image input, ground-truth UV map, masked UV input; The right side displays the results generated from different models with different patch sizes, where from top to bottom: output UV map,  attention map, mapped back face and the highlighted regions.
}
\label{fig-patch-size}
\end{figure*}

%% file: tables/vd_speed.tex
\begin{table}[t]
\caption{Inference speed of video inpainting methods.
}
\label{table:vd-speed}
\vspace{-1em}
\begin{center}
\setlength{\tabcolsep}{3pt}
\resizebox{\linewidth}{!}{
\begin{tabu}{p{1.3cm}|
*{4}{|@{}p{2cm}<{\centering}@{}}
}
\noalign{\hrule height 1pt}
Method & 3DGated~\cite{3dconv-vd} & STTN~\cite{sttn} &   FuseFormer \cite{liu2021fuseformer} & Ours  \\

 \hline 
Speed (fps) 
&  15.7 &  22.4    &  26.1  &  19.3 
       \Bstrut\\
\noalign{\hrule height 1pt}
\end{tabu}
} 
\end{center}
\end{table}

%% file: imgs/fig-user-study.tex
\begin{figure}[t]
\begin{center}
  \includegraphics[width=\linewidth]{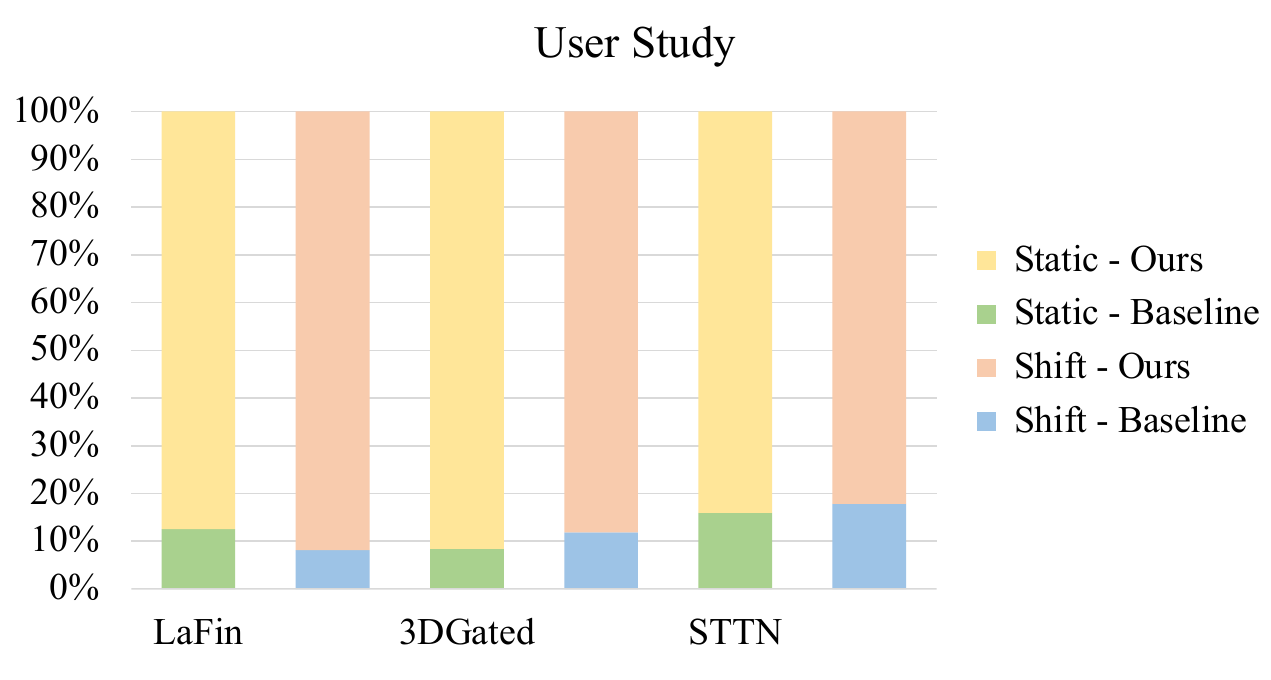}
\end{center}
\caption{User study results. We test on both static mask and shifting mask to evaluate the visual quality of our method. The test is conducted in one-to-one manner. For each baseline, left column is the result of static mask, and right column is for shifting mask. For each column, the upper stack shows the preference of our method, while the bottom stack shows the percentage for each baseline.
}
\label{fig-user-study}
\end{figure}

%% file: imgs/fig-application.tex
\begin{figure}[t]
\begin{center}
  \includegraphics[width=\linewidth]{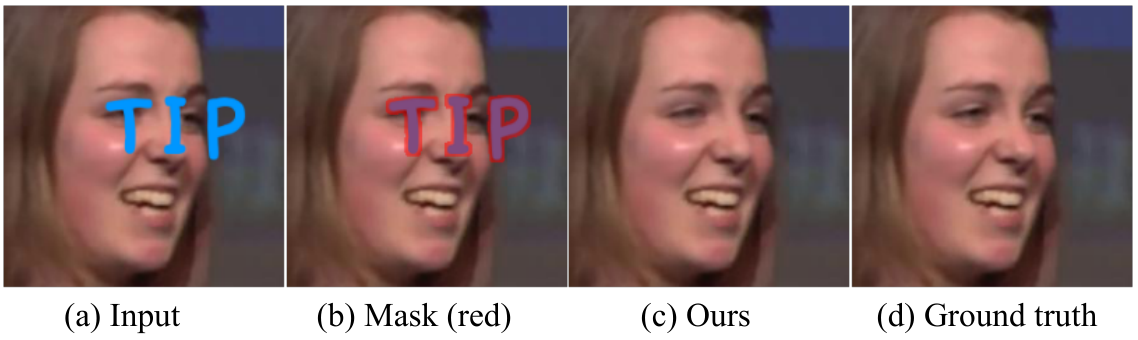}
\end{center}
\caption{Example of video watermark removal. 
}
\label{fig-application}
\end{figure}

%% file: imgs/fig-failure.tex
\begin{figure}[t]
\begin{center}
  \includegraphics[width=\linewidth]{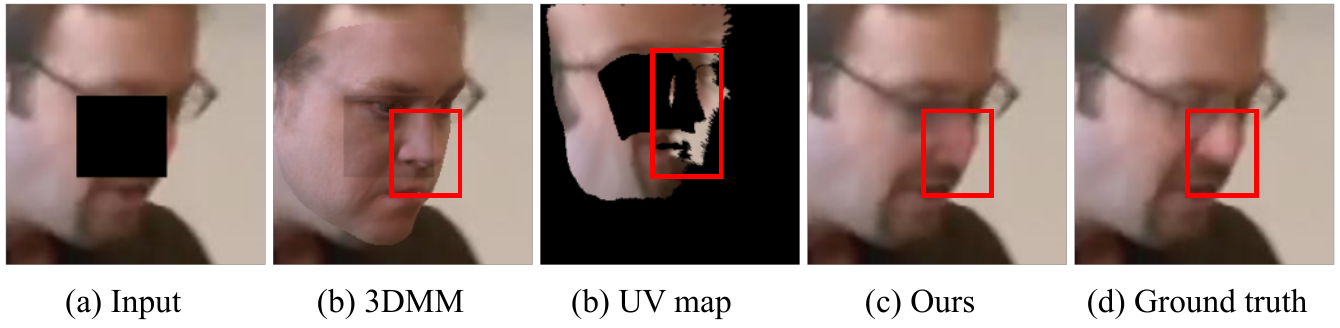}
\end{center}
\caption{Failure cases of our method. 
}
\label{fig-failure}
\end{figure}

%% file: text/5_conclusion.tex

\section{Conclusion}
In this paper, we propose a novel approach to facilitate face video inpainting by exploring face texture completion in the UV space.
The symmetry and aligned distribution of face textures in the UV space help to restore the masked regions with detailed face textures and structures.
We design a \uvmodel\ with a \uvattn\ module to enable efficient frame-wise correspondence retrieval from reference UV texture maps.
Compared with existing state-of-the-art methods, our approach is capable of synthesizing more visually plausible results especially under large face pose and expression variations. 